\documentclass{article}

\usepackage{spconf,amsmath,graphicx} 
\usepackage{amssymb}
\usepackage{bm}
\usepackage[T1]{fontenc}
\usepackage{subfigure}
\usepackage{booktabs}
\usepackage[table]{xcolor}
\usepackage{multirow}
\usepackage{siunitx}
\usepackage{float}
\usepackage[switch]{lineno}

\graphicspath{{./figures/}{./figs/result/}}

\usepackage[colorlinks,
            citecolor=red,
            urlcolor=red,
            linkcolor=red,
            bookmarks=false,
            hypertexnames=true]{hyperref}

\title{Floating-Body Hydrodynamic Neural Networks}

\name{
  Tianshuo Zhang \qquad
  Wenzhe Zhai \qquad
  Rui Yann \qquad 
  Jia Gao \qquad
  He Cao \qquad
  Xianglei Xing$^\ast$\thanks{$^\ast$Corresponding author: xingxl@hrbeu.edu.cn.} 
}
\address{Harbin Engineering University}

\begin{document}
\ninept

\maketitle


\begin{abstract}
Fluid-structure interaction is common in engineering and natural systems, where floating-body motion is governed by added mass, drag, and background flows. Modeling these dissipative dynamics is difficult: black-box neural models regress state derivatives with limited interpretability and unstable long-horizon predictions.
We propose Floating-Body Hydrodynamic Neural Networks (FHNN), a physics-structured framework that predicts interpretable hydrodynamic parameters such as directional added masses, drag coefficients, and a streamfunction-based flow, and couples them with analytic equations of motion. This design constrains the hypothesis space, enhances interpretability, and stabilizes integration.
On synthetic vortex datasets, FHNN achieves up to an order-of-magnitude lower error than Neural ODEs, recovers physically consistent flow fields. Compared with Hamiltonian and Lagrangian neural networks, FHNN more effectively handles dissipative dynamics while preserving interpretability, which bridges the gap between black-box learning and transparent system identification.
\end{abstract}


\section{Introduction}
\label{sec:intro}

Modeling dynamical systems from observed trajectories is a central challenge in signal processing, control, and scientific machine learning~\cite{brunton2022data, champion2019data, stiasny2021physics}. 
Classical approaches estimate governing parameters from noisy observations, whereas modern neural methods such as Neural Ordinary Differential Equations (Neural ODEs)~\cite{chen2018neural}, Hamiltonian Neural Networks (HNNs)~\cite{greydanus2019hamiltonian}, and Lagrangian Neural Networks (LNNs)~\cite{cranmer2020lagrangian} leverage differentiable solvers to directly learn vector fields or energy functions from data, enabling flexible modeling of complex dynamics~\cite{chensymplectic}.

\begin{figure}[htb]
  \centering
  \includegraphics[width=0.8\linewidth]{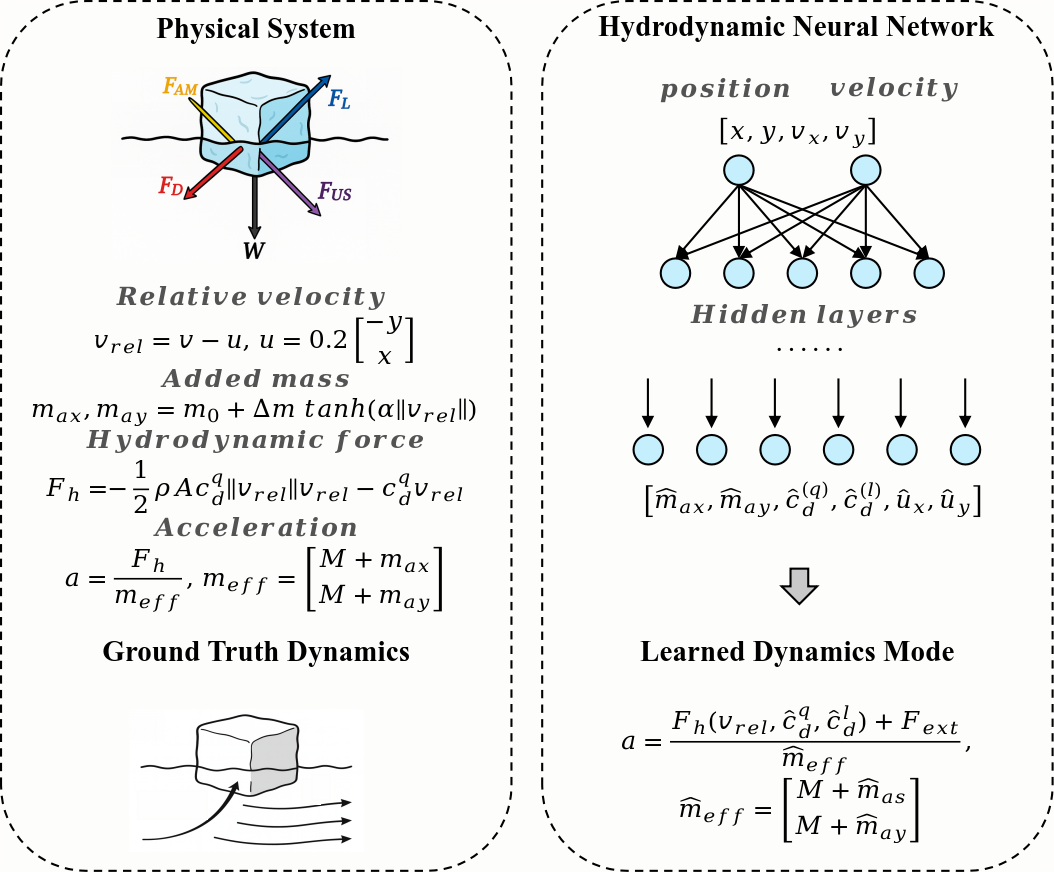}
  \caption{The Framework of Hydrodynamic Neural Network.}
  \label{fig:Framework}
\end{figure}

Despite their progress, existing neural approaches face two limitations. 
Black-box models that regress state derivatives offer limited interpretability and obscure links to physical quantities such as mass, drag, or flow velocity~\cite{wang2024identifiability, giampiccolo2024robust}. 
Energy-based formulations (HNNs and LNNs) are effective for conservative systems but struggle to incorporate dissipation, including drag forces and unsteady background flows. 
Consequently, long-horizon rollouts often degrade and the learned dynamics lack physical transparency~\cite{sundararaghavan2024lagrangian}.

In many real scenarios, especially fluid--structure interaction, dynamics are dissipative and governed by interpretable hydrodynamic parameters~\cite{francis2024modelling, yu2024learning}.
Added mass effects, linear and quadratic drag, and background flow determine the motion of floating bodies~\cite{chu2024viscous}. 
Accurate recovery of these quantities from trajectory data enables both stable long-term prediction and physically meaningful interpretation.

To address these challenges, we introduce \textbf{Floating-Body Hydrodynamic Neural Networks (FHNN)}, a physics-structured framework for hydrodynamic system identification, as illustrated in Fig.~\ref{fig:Framework}.
Instead of directly predicting accelerations, FHNN estimates directional added masses, linear and quadratic drag coefficients, and a streamfunction-based flow, which are then integrated into analytic equations of motion to compute accelerations.
This hybrid formulation constrains the hypothesis space, enhances interpretability, and improves stability in long-horizon integration.

Our contributions are threefold.
First, we propose FHNN, a physics-structured architecture that predicts hydrodynamic parameters (added mass, drag, and background flow) rather than raw derivatives, directly linking learning to interpretable physical quantities. 
Second, FHNN recovers these parameters and flow fields, yielding physically consistent insights into floating-body dynamics beyond conventional neural models. 
Third, on synthetic vortex datasets with trajectory-level splits and long-horizon evaluation, FHNN achieves lower errors, stable rollouts, and consistent parameter estimates, outperforming both black-box Neural ODEs and mainstream physics-guided networks.
Overall, FHNN establishes a physics-aware parameterization that connects black-box learning with interpretable identification of floating-body dynamics, offering broad potential for dissipative dynamical systems.

\section{Theory}
\label{sec:theory}

We model the dynamics of a rigid body translating in a two-dimensional incompressible background flow.  
The state vector is
\begin{equation}
s=(x,\,y,\,v_x,\,v_y)^\top,
\label{eq:state}
\end{equation}
where $\boldsymbol{x}=(x,y)$ denotes the position and $\boldsymbol{v}=(v_x,v_y)$ the velocity of the body.  

\subsection{Background flow}  
The surrounding fluid is described by a divergence-free velocity field $\boldsymbol{u}(\boldsymbol{x})$, represented compactly via a scalar stream function $\psi(x,y)$:
\begin{equation}
\boldsymbol{u}(\boldsymbol{x})=\nabla^\perp\psi(\boldsymbol{x}),
\label{eq:streamfunc}
\end{equation}
which by construction satisfies $\nabla\!\cdot\!\boldsymbol{u}=0$.  
This formulation avoids explicitly enforcing incompressibility, since it holds identically.

\subsection{Relative velocity}  
The dynamics depend on the relative velocity of the body with respect to the background flow,
\begin{equation}
\boldsymbol{v}_{\mathrm{rel}}=\boldsymbol{v}-\boldsymbol{u}(\boldsymbol{x}),\qquad
\sigma=\|\boldsymbol{v}_{\mathrm{rel}}\|+\varepsilon,
\label{eq:vrel}
\end{equation}
where the small constant $\varepsilon$ prevents degeneracy when the body moves exactly with the flow.  
Here $\sigma$ acts as a scalar measure of the effective Reynolds number scale.

\subsection{Fluid forces}  
Newton's second law gives
\begin{equation}
M\dot{\boldsymbol{v}}=\boldsymbol{F}_h+\boldsymbol{F}_{\mathrm{ext}},
\label{eq:newton}
\end{equation}
where $M$ is the dry mass of the body, $\boldsymbol{F}_{\mathrm{ext}}$ denotes external forces (e.g., gravity or actuation), and $\boldsymbol{F}_h$ the hydrodynamic force.  

\subsection{Added mass}  
A rigid body accelerating in a fluid must also accelerate the displaced fluid, which can be captured through an anisotropic added mass model \cite{wang2024numerical}.  
The effective mass matrix is
\begin{equation}
M_{\mathrm{eff}}=
\begin{bmatrix}
M+m_{ax} & 0 \\[2pt]
0 & M+m_{ay}
\end{bmatrix},
\label{eq:meff}
\end{equation}
where $m_{ax}$ and $m_{ay}$ are the added masses along the $x$- and $y$-directions.  

\subsection{Drag forces}  
Viscous and pressure drag are modeled through quadratic and linear terms in $\boldsymbol{v}_{\mathrm{rel}}$:
\begin{equation}
\boldsymbol{F}_q = -\tfrac12\rho A c_q\,\sigma\,\boldsymbol{v}_{\mathrm{rel}},\quad
\boldsymbol{F}_l = -c_l\,\boldsymbol{v}_{\mathrm{rel}},\quad
\boldsymbol{F}_h = \boldsymbol{F}_q+\boldsymbol{F}_l.
\label{eq:drag}
\end{equation}
Here $\rho$ is the fluid density, $A$ a reference area, and $c_q,c_l\!\geq\!0$ are empirical drag coefficients.  

\subsection{Equations of motion}  
Collecting the above, the governing dynamics of the system can be written as
\begin{equation}
\dot s=
\begin{bmatrix}
v_x \\ v_y \\ a_x \\ a_y
\end{bmatrix},\qquad
\boldsymbol{a}=M_{\mathrm{eff}}^{-1}
(\boldsymbol{F}_h+\boldsymbol{F}_{\mathrm{ext}}),
\label{eq:dynamics}
\end{equation}
where $\boldsymbol{a}$ denotes the acceleration of the rigid body.  
This formulation clearly separates the effects of added mass, drag, and external forcing, while ensuring consistency with incompressible background flow.

\section{Related Work}
\label{sec:related}

\subsection{Physics Priors}
A long line of research has incorporated physical inductive biases into neural models of dynamical systems. 
Early methods enforced invariants such as energy or momentum conservation by constraining the function class \cite{greydanus2019hamiltonian,cranmer2020lagrangian} or penalizing violations of physical laws, while later work introduced coordinate- or symmetry-based structure, \emph{e.g.,} parameterizing vector fields as potential gradients or leveraging Euclidean equivariance \cite{finzi2020generalizing,bronstein2021geometric}. 
In fluid-structure interaction, physics priors have been applied to guarantee incompressibility \cite{raissi2019physics}, enforce non-negative dissipation, and ensure frame-invariant force parameterizations, thereby reducing sample complexity and improving interpretability in scientific machine learning.

\subsection{Physics-guided Networks}
Our work relates to neural architectures that learn dynamics from physical principles. HNN~\cite{greydanus2019hamiltonian} encodes a Hamiltonian to enforce symplectic structure and energy conservation, while LNN~\cite{cranmer2020lagrangian} parameterizes a Lagrangian and derives motion via the Euler-Lagrange equations, ensuring coordinate invariance. 
Extensions have incorporated dissipation, control, and scaling to larger systems. 
More recently, Denoising Hamiltonian Networks (DHNs)~\cite{deng2025denoising} generalize Hamiltonian operators with denoising objectives and global conditioning, improving stability and multi-system modeling. 
Unlike these methods, which mainly target rigid-body dynamics in finite-dimensional coordinates, our formulation addresses hydrodynamic interactions in continuous flows, embedding priors such as incompressibility, added mass, and drag. 
In this sense, our model is a fluid-dynamical analogue of structured networks, tailored for fluid-structure coupling.

\section{Methodology}
\label{sec:method}

\subsection{Rotation-invariant parameterization}  
To ensure that the hydrodynamic model respects rotational symmetries, we do not directly learn direction-dependent coefficients.  
Instead, a neural coefficient network takes as input the radial distance
\begin{equation}
r=\|\boldsymbol{x}\|,
\end{equation}
and the relative speed $\sigma$ from (\ref{eq:vrel}), and outputs
\[
\{\,m_{ax}(r,\sigma),\ m_{ay}(r,\sigma),\ c_q(r,\sigma),\ c_l(r,\sigma)\,\}.
\]
These quantities enter the effective mass (\ref{eq:meff}) and drag formulation (\ref{eq:drag}), thus shaping the dynamics in (\ref{eq:dynamics}).  
This design makes the learned coefficients depend only on rotation-invariant scalars $(r,\sigma)$, rather than raw coordinates, leading to better generalization.

\subsection{Parameter constraints}  
To guarantee physical plausibility, all coefficients are constrained to be nonnegative and bounded.  
Given a raw network output $z$, we apply
\begin{equation}
\hat z=C\cdot\sigma(\mathrm{softplus}(z)/C),
\end{equation}
where $C=\{60,60,2,10\}$ for $\{m_{ax},m_{ay},c_q,c_l\}$ respectively.  
This mapping softly caps each parameter, preventing divergence while retaining smooth differentiability for optimization.

\subsection{Loss functions}  
The learning objective consists of three complementary terms:
\begin{itemize}
    \item \emph{Derivative loss:} Encourages the learned dynamics $f_\theta$ to match instantaneous accelerations derived from ground truth,
    \begin{equation}
    \mathcal{L}_{\mathrm{deriv}}
    =\mathbb{E}_{(s,\dot s^\star)}\|f_\theta(s,0)-\dot s^\star\|_2^2.
    \end{equation}

    \item \emph{Step loss:} Matches finite-time rollouts by comparing the RK4-integrated state $\Phi_\theta^\Delta(s)$ with the observed next state $s^+$,
    \begin{equation}
    \mathcal{L}_{\mathrm{step}}
    =\mathbb{E}_{(s,s^+)}\|\Phi_\theta^\Delta(s)-s^+\|_2^2.
    \end{equation}

    \item \emph{Flow smoothness:} Regularizes the background flow $\boldsymbol{u}$ in (\ref{eq:streamfunc}) by penalizing the Frobenius norm of the Hessian of $\psi$,
    \begin{equation}
    \mathcal{L}_{\mathrm{smooth}}
    =\lambda_{\mathrm{flow}}\|H\psi\|_F^2.
    \end{equation}
    This encourages physically coherent, divergence-free flows rather than noisy or oscillatory velocity fields.
\end{itemize}

The overall training objective is
\begin{equation}
\mathcal{L}=\mathcal{L}_{\mathrm{deriv}}+\mathcal{L}_{\mathrm{step}}+\mathcal{L}_{\mathrm{smooth}}.
\end{equation}

\subsection{Training protocol}  
Synthetic trajectories are split strictly at the trajectory level to avoid leakage across train/test sets.  
We optimize using Adam with a piecewise-decaying learning rate schedule.  
For long-horizon evaluation, we integrate $f_\theta$ with a high-accuracy ODE solver and compute mean squared error (MSE) of both positions $\boldsymbol{x}$ and velocities $\boldsymbol{v}$ at integer-second checkpoints.  


\section{Experiments}
\label{sec:experiments}

We evaluate the proposed FHNN on synthetic fluid-structure interaction trajectories and compare against black-box baselines and mainstream Physics-guided Networks. 
Performance is assessed using four metrics: mean-square error (MSE), root mean squared error (RMSE), average displacement error (ADE), and final displacement error (FDE)~\cite{mohamed2022social,phong2023truly}. 
The analysis focuses on qualitative prediction quality, flow field recovery, long-horizon stability, and the role of physics priors and network design.

\begin{figure}[ht]
  \centering
  \subfigure[]{\includegraphics[width=0.11\textwidth]{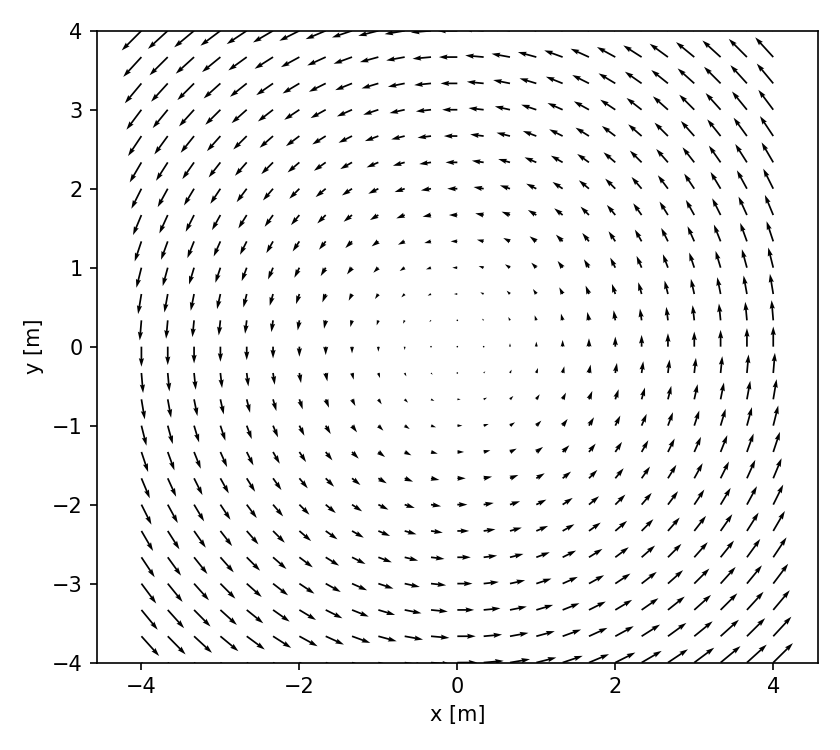}}
  \subfigure[]{\includegraphics[width=0.11\textwidth]{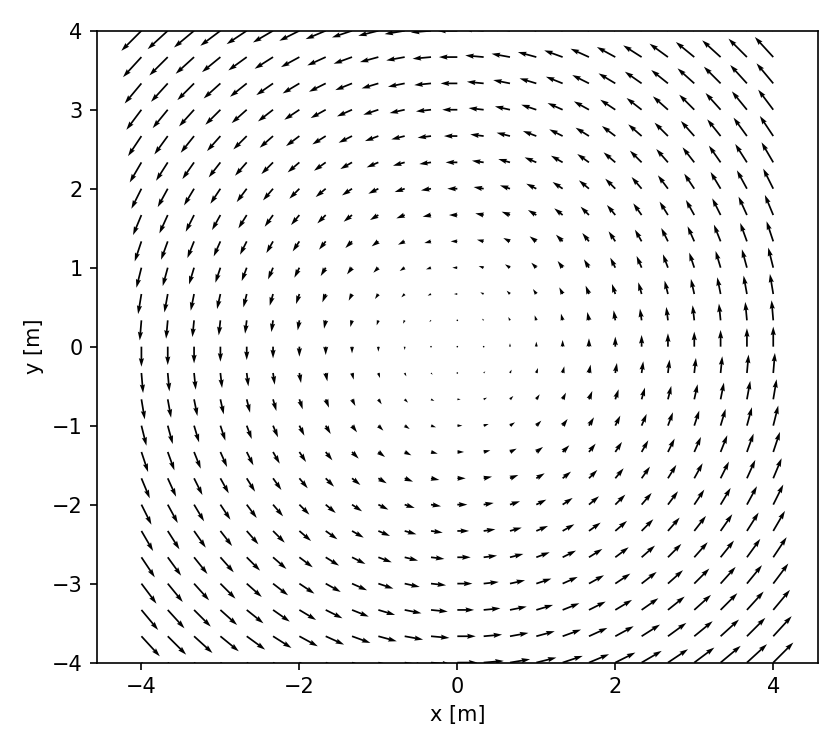}}
  \subfigure[]{\includegraphics[width=0.11\textwidth]{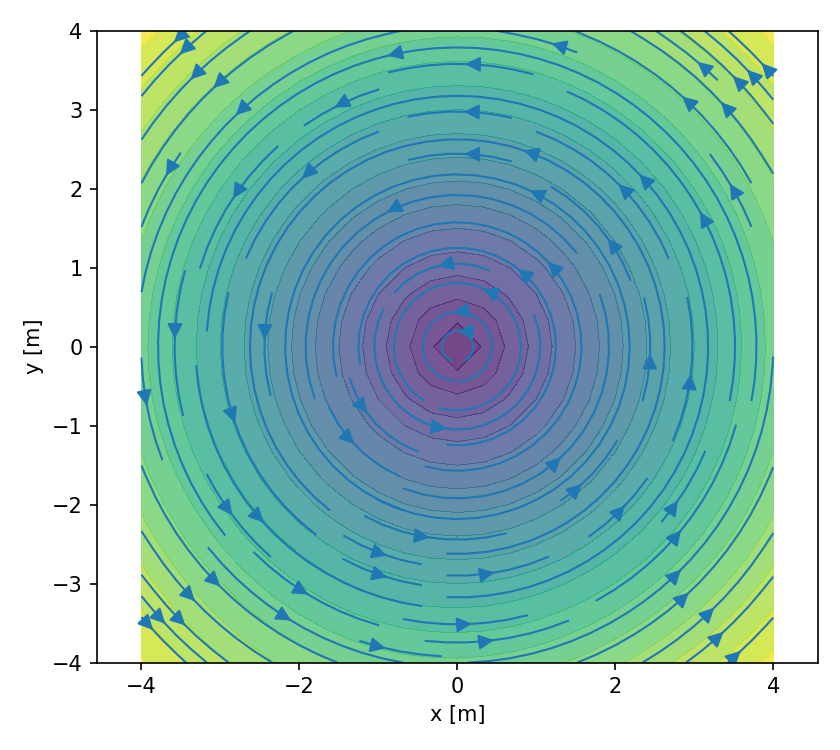}}
  \subfigure[]{\includegraphics[width=0.11\textwidth]{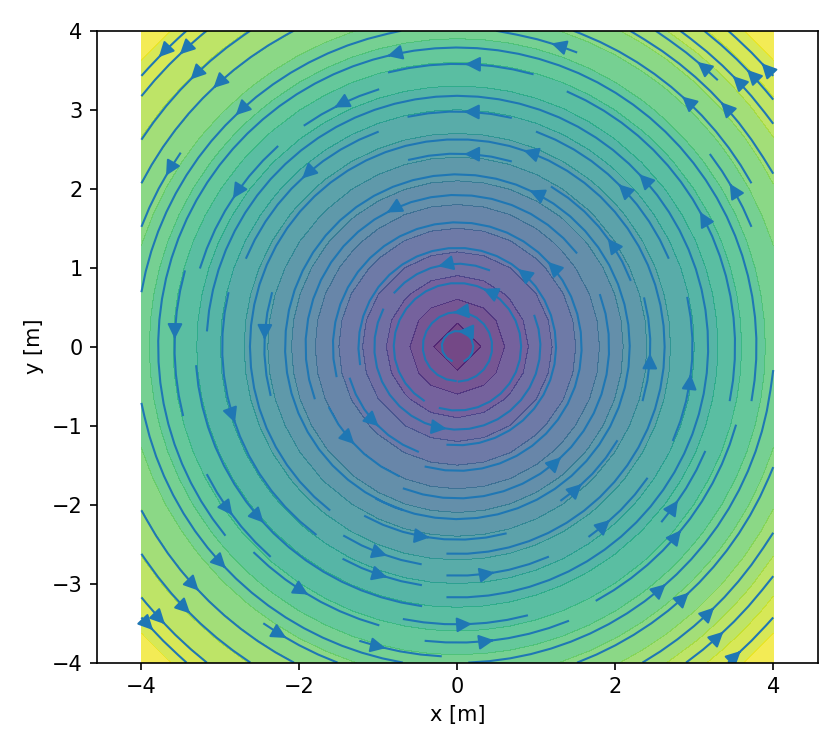}}
  \caption{Learned vs. true flow field: quiver(a,b) / stream(c,d).}
  \label{Flow Field}
\end{figure}

\subsection{Flow Field Recovery} 
Fig.~\ref{Flow Field} compares the learned streamfunction induced flow field against the true field on the same spatial grid. 
Subfigures (a) and (b) show that the model recovers an almost identical counter-clockwise vortex and azimuthal velocity distribution. 
Subfigures (c) and (d) further demonstrate close agreement in streamline topology and in the monotonic increase of speed magnitude with radius, while preserving incompressibility and rotational symmetry. 
Apart from minor edge effects due to arrow density, the two fields are qualitatively consistent and quantitatively close in direction, intensity, and structure, confirming the physical interpretability of the learned streamfunction.

\begin{figure}[ht]
  \centering
  \includegraphics[width=\linewidth]{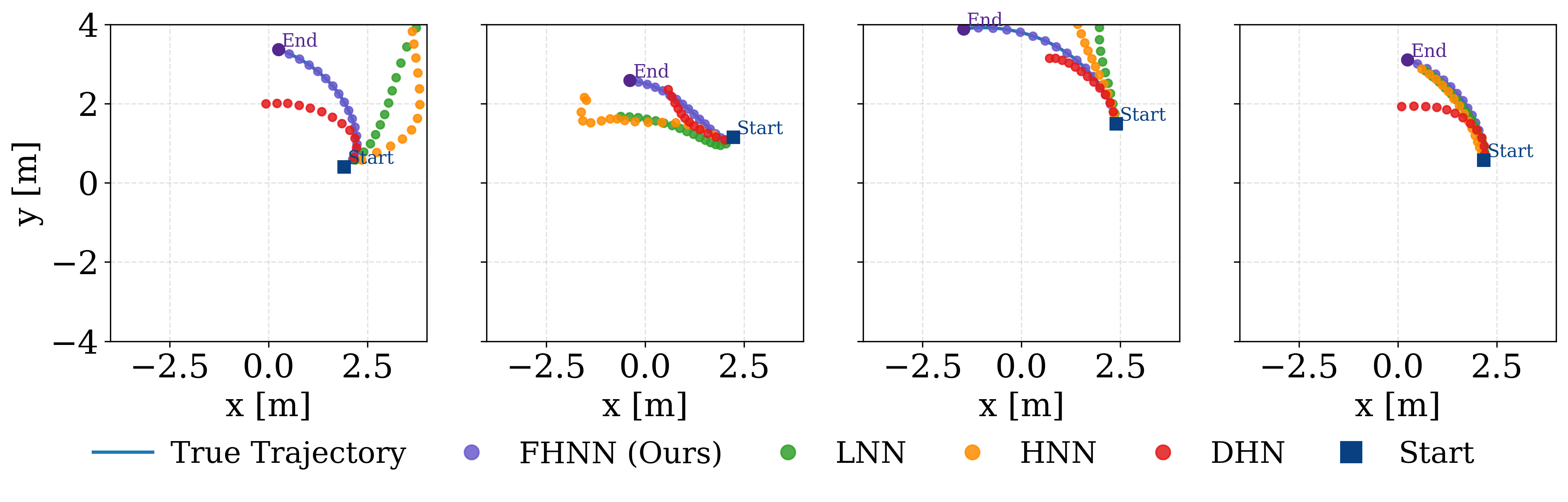}
  \caption{Example trajectory rollouts under different models.}
  \label{fig:traj_examples}
\end{figure}

\subsection{Trajectory Prediction} 
As shown in Fig.~\ref{fig:traj_examples}, we conduct a qualitative evaluation on four unseen 
initial conditions. 
The physically structured network accurately reproduces the ground-truth trajectories: the predicted samples nearly overlap the ground-truth curves with negligible phase drift, and the start and end positions align well. 
The model stably captures the rotation-induced curved paths, exhibiting only minor deviations even at high-curvature turns. 
These results indicate that the learned streamfunction and force coefficients generalize effectively and remain numerically stable under long-horizon integration.

\subsection{Comparison with Neural ODE} 
Table~\ref{tab:longhorizon_comparison} reports the long-horizon evaluation of FHNN compared with a black-box Neural ODE~\cite{chen2018neural}, averaged over 10 test 
trajectories. Metrics include RMSE of position and velocity, as well as FDE. 
FHNN outperforms the Neural ODE at both 8~s and 16~s horizons, achieving an order of 
magnitude lower errors and maintaining stable rollouts. This highlights the 
effectiveness of incorporating physics priors compared to purely black-box approaches.

\begin{table}[ht]
  \centering
  \caption{Evaluation of FHNN vs. Neural ODE.}
  \label{tab:longhorizon_comparison}
  \resizebox{\columnwidth}{!}{%
  \begin{tabular}{l S S S S S S}
    \toprule
    \textbf{Time} &
    \multicolumn{3}{c}{\textbf{Floating-Body HNN}} &
    \multicolumn{3}{c}{\textbf{Black-box Neural ODE}} \\
    \cmidrule(lr){2-4}\cmidrule(lr){5-7}
     & {\textbf{RMSE(pos) [m]}} & {\textbf{RMSE(vel) [m/s]}} & {\textbf{FDE(pos) [m]}}
     & {\textbf{RMSE(pos) [m]}} & {\textbf{RMSE(vel) [m/s]}} & {\textbf{FDE(pos) [m]}} \\
    \midrule
    1s  &
    \cellcolor{gray!15}\bfseries \num{1.07e-04} &
    \cellcolor{gray!15}\bfseries \num{1.52e-04} &
    \cellcolor{gray!15}\bfseries \num{2.13e-04} &
    \num{4.47e-03} & \num{4.11e-03} & \num{8.94e-03} \\
    2s &
    \cellcolor{gray!15}\bfseries \num{2.36e-04} &
    \cellcolor{gray!15}\bfseries \num{1.86e-04} &
    \cellcolor{gray!15}\bfseries \num{5.37e-03} &
    \num{7.38e-03} & \num{4.94e-03} & \num{1.56e-02} \\
    3s  &
    \cellcolor{gray!15}\bfseries \num{3.67e-04} &
    \cellcolor{gray!15}\bfseries \num{4.02e-04} &
    \cellcolor{gray!15}\bfseries \num{5.37e-04} &
    \num{1.00e-02} & \num{5.72e-03} & \num{2.15e-02} \\
    4s &
    \cellcolor{gray!15}\bfseries \num{4.91e-04} &
    \cellcolor{gray!15}\bfseries \num{2.10e-04} &
    \cellcolor{gray!15}\bfseries \num{1.15e-04} &
    \num{1.27e-02} & \num{6.13e-03} & \num{2.76e-02} \\
    \bottomrule
  \end{tabular}}
  
\end{table}

\subsection{Comparison with mainstream Physics-guided Networks}
\begin{figure}[ht]
  \centering
  \includegraphics[width=\linewidth]{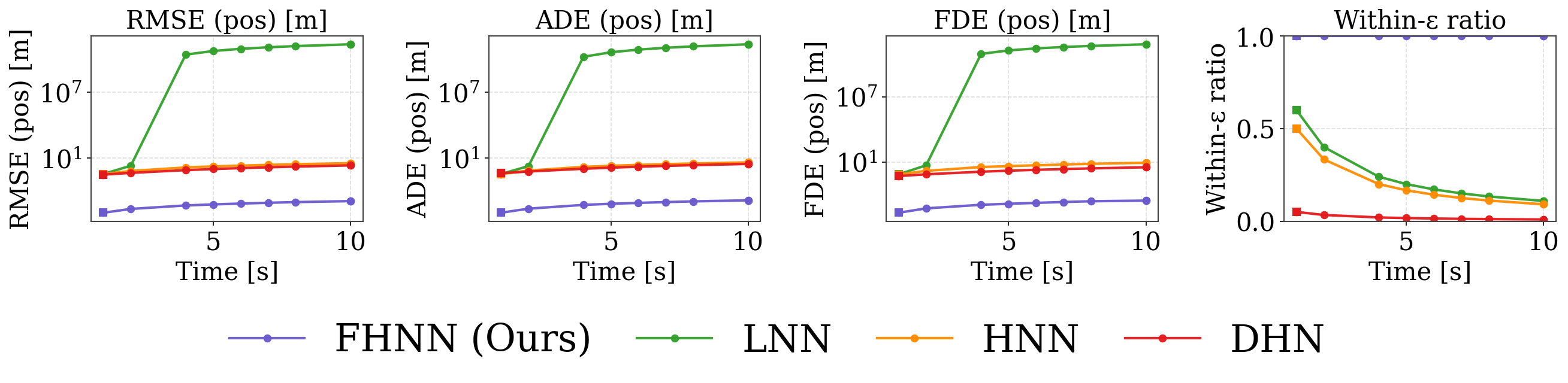}
  \caption{Multi-horizon position errors.}
  \label{fig:metrics_vs_horizon}
\end{figure}

We benchmark FHNN against representative physics-guided baselines: HNNs~\cite{greydanus2019hamiltonian}, LNNs~\cite{cranmer2020lagrangian}, and DHNs~\cite{deng2025denoising}. 
Quantitative results are reported in Table~\ref{tab:Trajectory prediction errors}, while Fig.~\ref{fig:traj_examples} and Fig.~\ref{fig:metrics_vs_horizon} provide multi-horizon metrics and trajectory visualizations.

At short horizons (1\,s), all baselines achieve errors on the order of $10^{-1}$, whereas FHNN reduces them by three orders of magnitude ($10^{-4}$). 
At longer horizons (5\,s), the gap widens significantly: LNN diverges catastrophically ($10^{10}$--$10^{11}$ errors), HNN accumulates multi-meter drift, and DHN remains unstable with errors near unity. 
In contrast, FHNN sustains errors below $10^{-3}$ across RMSE, ADE, and FDE, yielding up to eight orders of magnitude improvement over alternatives.

\begin{table}[ht]
  \centering
  \caption{Trajectory prediction errors: FHNN vs. baselines}
  \label{tab:Trajectory prediction errors}
  \resizebox{\columnwidth}{!}{%
    \begin{tabular}{l c S S S}
      \toprule
      \bfseries Time & \bfseries Model &
      {\bfseries RMSE} & {\bfseries ADE} & {\bfseries FDE} \\
      \midrule
      \multirow{4}{*}{1s}
        & HNN ~\cite{greydanus2019hamiltonian} & \num{3.54e-01} & \num{3.54e-01} & \num{7.08e-01} \\
        & LNN ~\cite{cranmer2020lagrangian} & \num{3.63e-01} & \num{3.63e-01} & \num{7.26e-01} \\
        & DHN ~\cite{deng2025denoising}   & \num{3.54e-01} & \num{3.54e-01} & \num{7.08e-01} \\
        & \cellcolor{gray!20}\textbf{FHNN}  & \cellcolor{gray!20}\bfseries \num{1.07e-04}
               & \cellcolor{gray!20}\bfseries \num{1.07e-04} & \cellcolor{gray!20}\bfseries \num{2.13e-04} \\
      \midrule
      \multirow{4}{*}{5s}
        & HNN ~\cite{greydanus2019hamiltonian} & \num{1.77e+00} & \num{2.03e+00} & \num{4.14e+00} \\
        & LNN ~\cite{cranmer2020lagrangian}& \num{5.85e+10} & \num{4.50e+10} & \num{1.83e+11} \\
        & DHN ~\cite{deng2025denoising}  & \num{1.02e+00} & \num{1.37e+00} & \num{1.57e+00} \\
        & \cellcolor{gray!20}\textbf{FHNN} & \cellcolor{gray!20}\bfseries \num{6.09e-04}
               & \cellcolor{gray!20}\bfseries \num{2.24e-04} & \cellcolor{gray!20}\bfseries \num{1.42e-03} \\
      \bottomrule
    \end{tabular}%
  }
\end{table}

Fig.~\ref{fig:metrics_vs_horizon} shows that FHNN remains stable up to 10\,s, while LNN diverges and HNN/DHN grow steadily. 
The within-$\varepsilon$ ratio stays near 1.0 for FHNN but drops to zero for baselines.
Fig.~\ref{fig:traj_examples} further confirms this gap: baselines overshoot or drift, whereas FHNN closely follows the ground truth. 
These results show that embedding incompressible flow constraints and rotation-invariant forces improves both short-term accuracy and long-horizon stability. 
The approach outperforms mainstream physics-guided networks by orders of magnitude.




\begin{table}[ht]
  \centering
  \caption{Long-horizon evaluation of FHNN.}
  \label{tab:longhorizon}
  \resizebox{\columnwidth}{!}{%
    \begin{tabular}{l
                    S[table-format=1.2e-2]
                    S[table-format=1.2e-2]
                    S[table-format=2.2e-2]}
      \toprule
      \textbf{Time} & \textbf{RMSE(pos) [m]} & \textbf{RMSE(vel) [m/s]} & \textbf{FDE(pos) [m]} \\
      \midrule
       30s  & \num{7.19e-02} & \num{2.02e-02} & \num{3.33e-01} \\
       60s  & \num{3.51e+00} & \num{6.62e-01} & \num{1.55e+01} \\
       120s & \num{1.98e+01} & \num{3.61e+00} & \num{3.87e+01} \\
      \bottomrule
    \end{tabular}%
  }
\end{table}

\begin{table}[ht]
  \centering
  \caption{Evaluation of FHNN across different flow scenarios.}
  \label{tab:fhnn_h5_scenarios}
  \resizebox{\columnwidth}{!}{%
    \begin{tabular}{l S S S S}
      \toprule
      \bfseries Scenario &
      {\bfseries RMSE(pos)} & {\bfseries RMSE(vel)} & {\bfseries ADE(pos)} & {\bfseries FDE(pos)} \\
      \midrule
      Noisy Flow           & \num{8.610e-03} & \num{2.355e-03} & \num{1.058e-02} & \num{1.779e-02} \\
      Flow with Obstacle   & \num{7.066e-03} & \num{1.994e-03} & \num{8.680e-03} & \num{1.442e-02} \\
      Steady Vortex        & \num{8.782e-03} & \num{2.405e-03} & \num{1.082e-02} & \num{1.801e-02} \\
      Time-varying Vortex  & \num{8.919e-03} & \num{2.516e-03} & \num{1.092e-02} & \num{1.869e-02} \\
      Wave (Morison Force) & \num{9.813e-03} & \num{3.016e-03} & \num{1.174e-02} & \num{2.156e-02} \\
      \bottomrule
    \end{tabular}%
  }
\end{table}

\subsection{Long-horizon Stability} 
To assess long-term behavior, Table~\ref{tab:longhorizon} reports errors up to 120~s. 
While the model remains highly accurate at 8--16~s, errors grow rapidly at longer horizons, with divergence evident at 60~s and beyond. 
This pattern reflects the challenge of compounding numerical errors in dissipative dynamics, even under a physics-structured design. 
Nonetheless, the stable short- to mid-term predictions are sufficient for many practical control and forecasting tasks.

\subsection{Scenario-wise Evaluation with Physical Insights} 
Table~\ref{tab:fhnn_h5_scenarios} reports FHNN accuracy at a 5~s horizon across representative flow regimes. 
The consistently low errors indicate that embedding fluid-structure priors constrains the dynamics effectively, even under stochastic noise or time-varying vorticity. 
The superior performance in the obstacle case highlights the model's ability to capture boundary-induced flow structures, while the larger errors in wave forcing reflect the inherent difficulty of oscillatory added-mass and drag effects. 
These results confirm that the physics-informed parameterization not only improves prediction accuracy but also yields interpretable performance variations aligned with known fluid mechanical principles.

\begin{table}[ht]
  \centering
  \caption{Ablation study of FHNN variants.}
  \label{tab:ablation}
  \resizebox{\columnwidth}{!}{%
    \begin{tabular}{l S S S S}
      \toprule
      \bfseries Model &
      {\bfseries RMSE(pos)} & {\bfseries RMSE(vel)} & {\bfseries ADE(pos)} & {\bfseries FDE(pos)} \\
      \midrule
      No Added Mass  & \num{3.35e-03} & \num{7.66e-04} & \num{4.12e-03} & \num{7.31e-03} \\
      No Linear Drag  & \num{2.85e-03} & \num{8.71e-04} & \num{3.17e-03} & \num{7.40e-03} \\
      No Flow Field  & \num{1.07e+00} & \num{2.62e-01} & \num{1.21e+00} & \num{2.71e+00} \\
      Shallow Net  & \num{1.29e-02} & \num{3.49e-03} & \num{1.48e-02} & \num{3.03e-02} \\
      ReLU  & \num{8.10e-01} & \num{1.94e-01} & \num{9.32e-01} & \num{2.02e+00} \\
      \cellcolor{gray!20}\textbf{FHNN}  & 
        \cellcolor{gray!20}\bfseries \num{9.92e-04} & 
        \cellcolor{gray!20}\bfseries \num{2.73e-04} &
        \cellcolor{gray!20}\bfseries \num{1.15e-03} & 
        \cellcolor{gray!20}\bfseries \num{2.42e-03} \\
      \bottomrule
    \end{tabular}%
  }
\end{table}

\begin{figure}[ht]
  \centering
  \includegraphics[width=\linewidth]{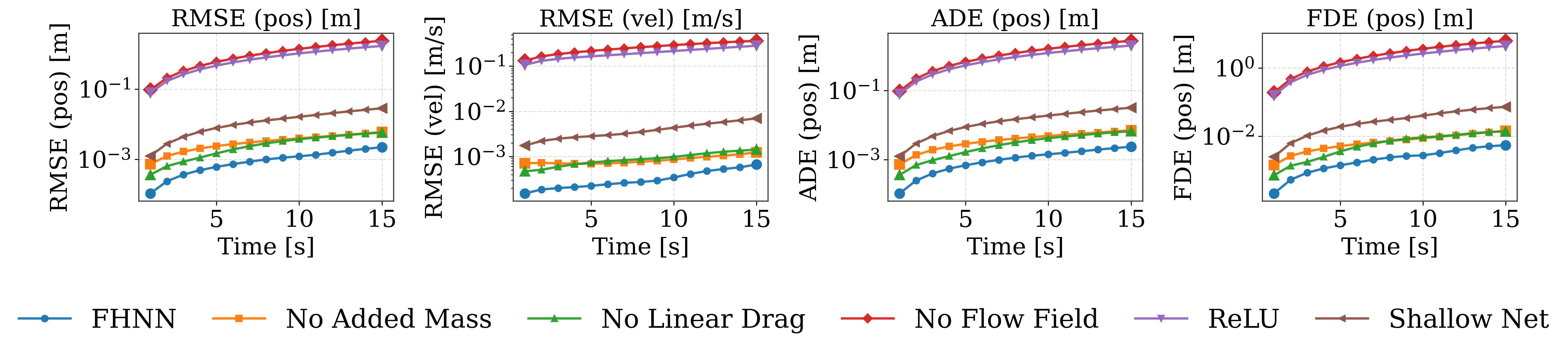}
  \caption{Performance of Ablated Models.}
  \label{fig:Ablated}
\end{figure}

\subsection{Ablation Study} 
Table~\ref{tab:ablation} and Fig.~\ref{fig:Ablated} present an ablation study to evaluate the importance of different physical priors and architectural choices. 
The baseline Floating-Body HNN achieves the best performance. 
Removing added mass or linear drag slightly degrades accuracy, while omitting the flow field causes catastrophic error growth and unstable rollouts, underscoring its critical role. Simplified network variants also lead to reduced accuracy and stability, confirming the need for smooth and expressive parameterizations. 
These results demonstrate that both physics-informed parameterization and adequate network capacity are essential for robust long-horizon modeling.

\section{Conclusion}
\label{sec:conclusion}

We presented a physics-structured neural framework for hydrodynamic system identification, designed to capture dissipative fluid-body interactions. 
The model predicts interpretable quantities, including added masses, drag coefficients, and a streamfunction-based flow field, and integrates them with analytic equations of motion. 
In doing so, it unites data-driven learning with physically grounded priors. 
This formulation improves interpretability, restricts the hypothesis space, and yields greater stability in long-horizon trajectory prediction. 
Looking forward, the framework opens several promising avenues. 
Extending the parameterization to three-dimensional flows and more complex body geometries could substantially broaden its scope. 
Incorporating real experimental measurements would allow testing beyond synthetic simulations.
Finally, integrating the approach with control objectives may enable interpretable and data-efficient strategies for robotic platforms operating in fluid environments.


\end{document}